\documentclass[letterpaper]{article} 
\usepackage{aaai2026}  
\usepackage{times}  
\usepackage{helvet}  
\usepackage{courier}  
\usepackage[hyphens]{url}  
\usepackage{graphicx} 
\usepackage{natbib}  
\usepackage{caption} 
\usepackage{amsfonts}
\usepackage{amsmath}
\usepackage{multirow}
\usepackage{booktabs}
\usepackage{xcolor}

\usepackage{tikz}
\usetikzlibrary{shapes, arrows}
\usepackage{algorithm}
\usepackage{algorithmic}

\usepackage{newfloat}
\usepackage{listings}
\DeclareCaptionStyle{ruled}{labelfont=normalfont,labelsep=colon,strut=off} 
\floatstyle{ruled}
\newfloat{listing}{tb}{lst}{}
\floatname{listing}{Listing}
\title{EmbedGrad: Gradient-Based Prompt Optimization in Embedding Space for Large Language Models}
\author{
    Xiaoming Hou\equalcontrib\textsuperscript{\rm 1, \rm 2},
    Jiquan Zhang\equalcontrib\textsuperscript{\rm 1},
    Zibin Lin\textsuperscript{\rm 1},
    DaCheng Tao\textsuperscript{\rm 3},
    Shengli Zhang\textsuperscript{\rm 1}\thanks{Corresponding author}
}
\affiliations{
    \textsuperscript{\rm 1}College of Electronics and Information Engineering, Shenzhen University\\
    \textsuperscript{\rm 2}College of Computer Science and Software Engineering, Shenzhen University\\
    \textsuperscript{\rm 3}Nanyang Technological University\\

}

\begin{document}

\maketitle

\begin{abstract}
Effectively adapting powerful pretrained foundation models to diverse tasks remains a key challenge in AI deployment. Current approaches primarily follow two paradigms: discrete optimization of text prompts through prompt engineering, or continuous adaptation via additional trainable parameters. Both exhibit limitations—discrete methods lack refinement precision while parameter-based techniques increase complexity and reduce interpretability.
To address these constraints, we propose EmbedGrad, a novel framework that optimizes text prompt embeddings through gradient-based refinement. Our approach uniquely decouples training from deployment: during optimization, labeled examples guide precise embedding adjustments while preserving semantic meaning; during inference, only optimized embeddings integrate with user queries. This enables fine-grained calibration impossible in text space, such as enhancing the reasoning capability of prompts like ``please reason step by step''.
Comprehensive evaluations across mathematical reasoning, sentiment analysis, and causal judgment tasks demonstrate EmbedGrad's effectiveness: optimizing this reasoning prompt for Qwen2.5-Math-1.5B increased accuracy from 14.74\% to 58.96\% on mathematical problems. Consistent improvements were observed across model scales (0.5B-14B) and all tasks, with particularly significant gains for smaller models on complex problems like causal judgment. By bridging prompt engineering and parameter efficiency without architectural changes, our work establishes embedding refinement as a powerful new paradigm for task adaptation.
\end{abstract}

\section{Introduction}
The impressive capabilities of pretrained foundation models have transformed artificial intelligence, but adapting these powerful systems effectively for real-world tasks remains challenging. Achieving strong performance on applications like mathematical reasoning, sentiment analysis, or causal judgment often requires careful adjustments, which typically need either labor-intensive prompt design or computationally expensive fine-tuning. This challenge is especially important for resource-limited environments where efficiency is critical.

Current adaptation approaches are divided into two main types, each with limitations. Natural language optimization methods work directly with text prompts, using techniques like AutoPrompt~\cite{autoprompt} and TextGrad~\cite{text_grad} to search for better wordings. While these methods produce understandable prompts, they struggle with language complexity and can only make coarse adjustments, often missing subtle meaning details. On the other hand, parameter-based methods such as LoRA~\cite{lora} and Prefix-Tuning~\cite{prefix_tuning} add task-specific components to the model, enabling precise control but increasing complexity, reducing transparency, and requiring more computation during use. Both approaches face a core tension: text optimization lacks precision, while parameter tuning sacrifices clarity and efficiency.

\begin{figure}[htbp]
\centering
\includegraphics[width=\linewidth]{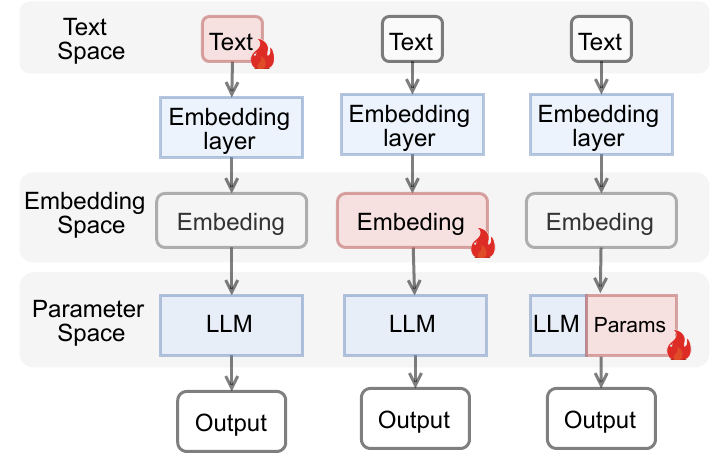}
\caption{Optimization Spaces: Text, Embedding (ours), and Parameter Domains.}
\label{fig_position}
\end{figure}

To solve this, we introduce EmbedGrad—a novel approach that bridges these methods by optimizing prompts in the continuous embedding space. As shown in Figure~\ref{fig_position} (middle), our method operates in the middle ground between text and parameter optimization, combining the understandability of natural language with the precision of continuous adjustments. EmbedGrad achieves this through three key innovations: (1) treating prompt embeddings as adjustable parameters that can be refined using gradients, (2) separating improvement from deployment to avoid extra inputs during actual use, and (3) working efficiently by only modifying prompt representations, not the model itself.

By working in the embedding space—where words become numerical representations—EmbedGrad enables precise meaning adjustments impossible through text changes. For example, refining the representation of ``please reason step by step'' increased math-solving accuracy by 44\% while keeping the prompt's original meaning. This approach allows fine-grained improvements along meaning pathways, such as strengthening the emotional focus of a sentiment prompt or enhancing causal reasoning capabilities without changing its words. Crucially, optimized prompts stay recognizable, with over 95\% similarity to the original wording.

Comprehensive tests across mathematical reasoning, sentiment analysis, and causal judgement tasks show EmbedGrad's effectiveness, with improved results across model sizes (0.5B to 14B parameters) and sustained efficiency with just 5-10 refinement cycles. These gains come from EmbedGrad's dual ability to make models more confident in reasoning while better activating key functions, confirmed through both confidence measurements and specialized analysis. The approach establishes a practical new way to adapt AI systems—delivering strong performance while keeping solutions understandable and efficient.

The remainder of this paper is structured as follows: Section~\ref{sec:rel} positions our work within existing literature; Section~\ref{sec:met} formalizes the EmbedGrad framework; Section~\ref{sec:exp} presents experimental validation; Section\ref{sec:dis} discusses mechanistic insights; and Section~\ref{sec:con} concludes with future directions.

\section{Related Work}\label{sec:rel}

\subsection{Natural Language Space Optimization} Prompt engineering in natural language space has evolved from manual refinement to automated techniques like AutoPrompt~\cite{autoprompt}, which generates discrete text templates through token-level optimization. While effective for tasks like sentiment analysis, it faces combinatorial complexity limitations. Self-refinement approaches such as Self-Refine~\cite{self_refine} employ LLM feedback loops to iteratively improve outputs, showing promise in mathematical reasoning but requiring multiple inference passes. TextGrad~\cite{text_grad} extends this paradigm by backpropagating natural language feedback. These methods share three core constraints: coarse-grained adjustments limited to token substitutions, unpredictable semantic drift from minor wording changes, and inference-phase redundancy where examples consume context window capacity. Crucially, they cannot leverage gradient-based optimization in continuous spaces.

\subsection{Parameter-Space Adaptation Methods} Parameter-efficient tuning augments models with learnable modules. Prompt Tuning~\cite{prompt_tuning} optimizes continuous ``soft prompts'' initialized randomly, achieving efficiency but sacrificing interpretability. Prefix-Tuning~\cite{prefix_tuning} inserts trainable vectors into transformer layers, while P-Tuning~\cite{p_tuning} and its variant~\cite{p_tuning_v2} generalizes prompt embeddings across layers and tasks—applying continuous prompts to each transformer layer rather than just the input. This enables stronger performance on NLU benchmarks but requires storing task-specific vectors. 
Adapter-based methods~\cite{adapter,adapter_drop,adapter_fusion} inserts bottleneck modules after Transformer blocks' self-attention and FFN layers, while LoRA-like methods~\cite{lora, qlora, adalora} implement low-rank adaptation by decomposing weight updates.
These approaches introduce architectural modifications and share critical limitations: optimization decouples from human-designed prompt semantics, and task-specific components increase deployment complexity.

\subsection{EmbedGrad's Distinct Positioning}  EmbedGrad fundamentally differs by bridging natural language and parameter spaces through embedding optimization of existing text prompts. Unlike Prompt Tuning that initializes random vectors, we preserve the semantic core of human-written prompts while enabling gradient-based refinement. Compared to Prefix-Tuning or P-Tuning v2 that insert layers of task parameters, EmbedGrad modifies only the prompt's embedding—requiring no architectural changes. Crucially, our approach uniquely decouples training (using examples) from inference (concatenating only optimized embeddings with queries), eliminating context window competition. This combines the interpretability of text prompts with the precision of continuous optimization, while maintaining parameter efficiency absent in adapter methods.

\section{Methodology}\label{sec:met}

\begin{figure*}[htbp]
\centering
\includegraphics[width=0.85\textwidth]{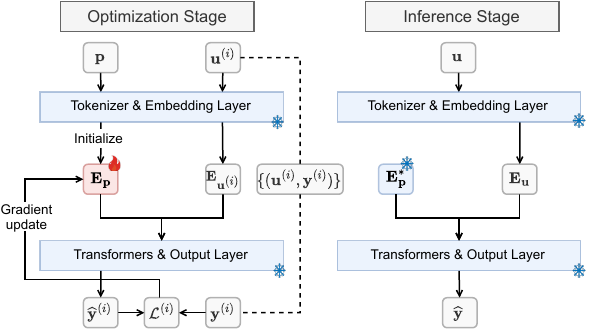}
\caption{The framework of EmbedGrad.}
\label{fig_framework}
\end{figure*}

\subsection{Preliminaries: Architecture of LLMs}
LLMs are typically autoregressive, generating outputs sequentially based on prior tokens. Their architecture consists of four core components:
\begin{enumerate}
    \item \textbf{Tokenizer}: Converts raw text into a sequence of tokens:
    \begin{equation}
        \mathbf{x} = [x_1,x_2, \cdots, x_n] \in \mathbb{N}^n.
    \end{equation}
    \item \textbf{Embedding Layer}: Maps tokens to continuous vector via an embedding matrix $ \mathbf{E} \in \mathbb{R}^{V \times d} $, yielding
    \begin{equation}
        \mathbf{e}_\mathbf{x} = \mathbf{E}_{\mathbf{x}} \in \mathbb{R}^{n \times d},
    \end{equation} where $V$ is the vocabulary size and $d$ is the embedding dimension. 
    \item \textbf{Transformer Layers}: Process embeddings autoregressively with masked attention. The hidden states $\mathbf{h}^{(L)}$ are computed layer-wise:
    \begin{equation}
            \mathbf{h}^{(i)} = 
            \begin{cases}
                \mathbf{e}_\mathbf{x}, &i=0 \\
                \text{Transformer}_{\theta_i}(\mathbf{h}^{(i-1)}), &i=1,2,\cdots,L \\
            \end{cases}
    \end{equation}
    Each position $t$ in $\mathbf{h}^{(L)}$ outputs a contextualized representation $\mathbf{h}_t^{(L)}$ for token $x_t$.
    \item \textbf{Output Layer}:  Predicts the next token at each step using:
    \begin{equation}
        \hat{\mathbf{y}}_t = \text{softmax}(W_o \mathbf{h}_t^{(L)} + \mathbf{b}_o),
    \end{equation}
    where $W_o \in \mathbb{R}^{d \times V}$ and $\mathbf{b}_o \in \mathbb{R}^V$. The final output $\hat{\mathbf{y}}$ is generated token-by-token autoregressively.
\end{enumerate}

\subsection{The EmbedGrad Framework}
EmbedGrad optimizes the continuous embedding of an existing natural language prompt while freezing all LLM parameters $\theta$, $W_o$, and $\mathbf{b}_o$. This framework comprises three stages: initialization, optimization, and inference, as depicted in Figure~\ref{fig_framework}.

\begin{itemize}
    \item \textbf{Initialization} A natural language prompt $P$ (e.g., ``Classify sentiment as positive/negative'') is tokenized into $\mathbf{p} = [p_1, p_2, \cdots, p_k]$. Its initial embedding $\mathbf{E}_\mathbf{p}^{(0)} \in \mathbb{R}^{k \times d}$ is directly extracted from the LLM's frozen embedding layer, preserving the original semantic structure.
    \item \textbf{Optimization} Given labeled data $\mathcal{D} = \{(\mathbf{u}^{(i)}, \mathbf{y}^{(i)})\}$, the concatenated prompt-and-input embedding becomes the input sequence:
    \begin{equation}
        \mathbf{h}^{(0)} = 
        \left[\mathbf{E}_\mathbf{p}, \mathbf{E}_{\mathbf{u}^{(i)}}\right] \in \mathbb{R}^{(k+n) \times d},
    \end{equation}
    where $\mathbf{E}_{\mathbf{u}^{(i)}}$ is the user input embedding.
    The LLM processes this through its autoregressive Transformer, computing hidden states $\mathbf{h}_t^{(L)}$ for each position $t$ in the concatenated sequence. The task loss $\mathcal{L}_{CE}$ is calculated over the output tokens $\hat{\mathbf{y}}_t$ and their targets $\mathbf{y}_t$, with updates propagated only to $\mathbf{E}_\mathbf{p}$:
    \begin{align}
        \mathcal{L}(\hat{\mathbf{y}}^{(i)}, \mathbf{y}^{(i)}) &= \sum_{t} \mathcal{L}_{CE}(\hat{\mathbf{y}}_t^{(i)}, \mathbf{y}_t^{(i)}), \\
        \mathbf{E}_{\mathbf{p}}^{*} &= \mathop{\arg\min}\limits_{\mathbf{E}_{\mathbf{p}}} \sum_{i=1}^{|\mathcal{D}|}\mathcal{L}(\hat{\mathbf{y}}^{(i)}, \mathbf{y}^{(i)}).
    \end{align}
    This aligns the prompt embedding with the autoregressive generation process without altering the model architecture.
    \item \textbf{Inference} The optimized embedding $\mathbf{E}_\mathbf{p}^*$ is concatenated with user input $\mathbf{u}$:
    \begin{equation}
        \text{Input} = [\mathbf{E}_\mathbf{p}^{*}, \mathbf{E}_{\mathbf{u}}] \in \mathbb{R}^{(k+m) \times d},
    \end{equation}
    where $m$ is the length of $\mathbf{u}$. The LLM then autoregressively generates output $\hat{\mathbf{y}}$ by sequentially attending to both the prompt and prior tokens:
    \begin{equation}
            \hat{\mathbf{y}} = \text{softmax}(W_o \cdot \text{Transformer}_\theta(\text{Input}) + \mathbf{b}_o)
    \end{equation}
    ensuring the prompt's semantic influence persists throughout the generation while avoiding redundant context usage.
\end{itemize}

\section{Experiments}\label{sec:exp}

\begin{table*}[htbp]
\centering
\caption{Initial prompts to be optimized with EmbedGrad for each dataset}
\label{tab:prompts}
\resizebox{0.8\textwidth}{!}{%
    \begin{tabular}{@{}lp{0.7\textwidth}@{}}
        \toprule
        \textbf{Dataset} & \textbf{Initial Prompt} \\
        \midrule
        Math500 & \texttt{please reason step by step, and put your final answer with \textbackslash{}boxed\{\}.} \\
        \addlinespace[2mm]
        IMDB & \texttt{Please determine whether the following statement is positive or negative. Let's think step by step.} \\
        \addlinespace[2mm]
        MEDD & \texttt{Determine the main emotion or mood expressed in the following text, choosing only from: [sad, angry, concerned, surprised, happy, calm, disgusted, doubtful]. Requirement: Let's think step by step and choose the strongest one.} \\
        \addlinespace[2mm]
        Bigbench & \texttt{How would a typical person answer each of the following questions about causation? Please think step by step, and then answer "Yes" or "No".}\\
        \bottomrule
    \end{tabular}%
}
\end{table*}

\begin{table*}[htbp]
\centering
\caption{Experimental results across tasks and models, with $\Delta$ showing absolute improvements.}
\label{tab:results}
\resizebox{0.8\textwidth}{!}{
    \begin{tabular}{@{}c c c c c c@{}}
        \toprule
        \textbf{Task} & \textbf{Dataset} & \textbf{Base Model} & \textbf{Before (\%)} & \textbf{After (\%)} & $\mathbf{\Delta}$ \textbf{(\%)} \\
        \midrule
        \multirow{12}{*}{\rotatebox[origin=c]{90}{Mathematical Reasoning}} 
        & \multirow{2}{*}{Math500} & Qwen2.5-Math-1.5B & 14.74 & 58.96 & \textcolor{green}{+44.22 $\uparrow$} \\
        & & Qwen2.5-Math-7B & 56.18 & 65.34 & \textcolor{green}{+9.16 $\uparrow$} \\
        \cmidrule(lr){2-6}
        & \multirow{2}{*}{Math500-L1} & Qwen2.5-Math-1.5B & 9.09 & 86.36 & \textcolor{green}{+77.27 $\uparrow$} \\
        & & Qwen2.5-Math-7B & 72.73 & 95.45 & \textcolor{green}{+22.72 $\uparrow$} \\
        \cmidrule(lr){2-6}
        & \multirow{2}{*}{Math500-L2} & Qwen2.5-Math-1.5B & 15.56 & 82.22 & \textcolor{green}{+66.66 $\uparrow$} \\
        & & Qwen2.5-Math-7B & 64.44 & 82.22 & \textcolor{green}{+17.78 $\uparrow$} \\
        \cmidrule(lr){2-6}
        & \multirow{2}{*}{Math500-L3} & Qwen2.5-Math-1.5B & 22.64 & 79.25 & \textcolor{green}{+56.61 $\uparrow$} \\
        & & Qwen2.5-Math-7B & 62.26 & 77.36 & \textcolor{green}{+15.10 $\uparrow$} \\
        \cmidrule(lr){2-6}
        & \multirow{2}{*}{Math500-L4} & Qwen2.5-Math-1.5B & 14.06 & 57.81 & \textcolor{green}{+43.75 $\uparrow$} \\
        & & Qwen2.5-Math-7B & 57.81 & 64.06 & \textcolor{green}{+6.25 $\uparrow$} \\
        \cmidrule(lr){2-6}
        & \multirow{2}{*}{Math500-L5} & Qwen2.5-Math-1.5B & 10.45 & 34.33 & \textcolor{green}{+23.88 $\uparrow$} \\
        & & Qwen2.5-Math-7B & 40.30 & 43.28 & \textcolor{green}{+2.98 $\uparrow$} \\
        \midrule
        \multirow{10}{*}{\rotatebox[origin=c]{90}{Sentiment Analysis}} 
        & \multirow{5}{*}{IMDB} & Qwen2.5-0.5B-Instruct & 20.24 & 34.52 & \textcolor{green}{+14.28 $\uparrow$} \\
        & & Qwen2.5-3B-Instruct & 59.52 & 71.43 & \textcolor{green}{+11.91 $\uparrow$} \\
        & & Qwen2.5-7B-Instruct & 69.05 & 78.57 & \textcolor{green}{+9.52 $\uparrow$} \\
        & & Llama-3.1-8B-Instruct & 57.14 & 64.29 & \textcolor{green}{+7.15 $\uparrow$} \\
        & & Qwen2.5-14B-Instruct & 70.24 & 77.38 & \textcolor{green}{+7.14 $\uparrow$} \\
        \cmidrule(lr){2-6}
        & \multirow{5}{*}{MEDD} & Qwen2.5-0.5B-Instruct & 16.03 & 90.57 & \textcolor{green}{+74.54 $\uparrow$} \\
        & & Qwen2.5-3B-Instruct & 66.37 & 93.41 & \textcolor{green}{+27.04 $\uparrow$} \\
        & & Qwen2.5-7B-Instruct & 87.36 & 92.10 & \textcolor{green}{+4.74 $\uparrow$} \\
        & & Llama-3.1-8B-Instruct & 83.57 & 92.10 & \textcolor{green}{+8.53 $\uparrow$} \\
        & & Qwen2.5-14B-Instruct & 84.67 & 92.90 & \textcolor{green}{+8.23 $\uparrow$} \\
        \midrule
        \multirow{4}{*}{\rotatebox[origin=c]{90}{\parbox{1.5cm}{\centering Causal\\Judgement}}}
        & \multirow{4}{*}{Bigbench} & Qwen2.5-0.5B-Instruct & 50.53 & 61.05 & \textcolor{green}{+10.58$\uparrow$} \\
        & & Qwen2.5-3B-Instruct & 54.74 & 63.16 & \textcolor{green}{+9.42$\uparrow$} \\
        & & Qwen2.5-7B-Instruct & 62.11 & 65.26 & \textcolor{green}{+3.15$\uparrow$} \\
        & & Qwen2.5-14B-Instruct & 69.47 & 71.58 & \textcolor{green}{+ 2.11$\uparrow$} \\ 
        \bottomrule
    \end{tabular}
}
\end{table*}

\subsection{Experimental Setup}
We conduct comprehensive evaluations to assess EmbedGrad's effectiveness across mathematical reasoning, sentiment analysis, and causal judgement tasks. For mathematical reasoning, we employ the Math500 benchmark~\cite{math500} containing 500 problems categorized into five difficulty levels (L1 to L5, with L1 representing elementary problems and L5 advanced challenges), which rigorously tests multi-step problem-solving capabilities. The sentiment analysis evaluation encompasses both binary classification using the English IMDB~\cite{imdb} movie review dataset and fine-grained 8-class emotion recognition with the Chinese MEDD dataset~\cite{medd}. For causal judgement, we use the BigBench benchmark~\cite{bigbench} which requires models to answer questions about causation. All experiments utilize greedy decoding, where at each generation step the highest-probability token is selected as the next output.

Our model selection spans diverse architectures and scales: Qwen2.5-Math-1.5B/7B for mathematical reasoning, Qwen2.5 instruct series (0.5B, 3B, 7B, and 14B parameters) and Llama-8B for sentiment analysis and causal judgement.
Critical optimization parameters include learning rates of 0.01 for 1.5B-scale models and 0.001 for larger models ($\geq$7B), training durations of 5-10 epochs with early stopping to prevent overfitting. The specific prompts optimized for each dataset are detailed in Table~\ref{tab:prompts}. These prompts were selected to represent common task formulations in their respective domains. During optimization, only the embedding representations of these text prompts were adjusted while preserving their semantic core.

\subsection{Experimental Results}

Comprehensive evaluation across diverse tasks demonstrates EmbedGrad's consistent effectiveness, as summarized in Table~\ref{tab:results}. The method delivers substantial accuracy improvements while eliminating inference-stage redundancy, with transformative results on complex tasks and smaller models. 

For mathematical reasoning, optimizing the prompt embedding elevated Qwen2.5-Math-1.5B accuracy from 14.74\% to 58.96\% (+44.22\%) on Math500, with dramatic gains on elementary problems (L1: +77.27\%). 
In sentiment analysis, optimizing task-specific prompts boosted Qwen-0.5B performance by 74.54\% on the challenging MEDD dataset. 
For causal judgement task on BigBench shows significant gains across model scales: Qwen-0.5B improved by 10.58\%, while larger models maintained consistent but more modest improvements. This reinforces our key observation that smaller models benefit most substantially from prompt optimization.

Our analysis reveals consistent patterns: improvements inversely correlate with model scale (Qwen-0.5B gains averaged 5.8× higher than Qwen-14B) and directly correlate with task complexity (MEDD improvements 3.2× higher than IMDB). Crucially, these benefits remain consistent across architectures and task types.
Beyond accuracy gains, EmbedGrad delivers efficiency advantages: optimization converges in 5-10 epochs—faster than fine-tuning—while reducing inference overhead by eliminating in-context examples. The most pronounced improvements occur where models struggle most: smaller architectures on complex tasks, demonstrating targeted enhancement without compromising semantic integrity.

\section{Discussion}\label{sec:dis}
\subsection{Semantic Anchoring in Continuous Optimization Space}
Our analysis of the optimized embedding vectors employs a rigorous token similarity assessment method to quantify semantic preservation. 
For each optimized token embedding $\mathbf{v}_{opt}$ in the prompt, we compute its similarity to the entire vocabulary through an inner product operation followed by softmax normalization: 
\begin{equation}
    p_j = \frac{\exp(\mathbf{v}_opt)\cdot\mathbf{e}_j}{\sum_{k=1}^{V}\exp(\mathbf{v}_opt)\cdot\mathbf{e}_j},
\end{equation}
where $\mathbf{e}_j$ represents embedding of all vocabulary tokens.
The token $w_{\hat{j}}$ with $\hat{j} = \mathop{\arg\max}_j p_j$ is identified as the nearest semantic match.
This quantitative approach revealed exceptionally high fidelity to the original prompt, with $p_{\text{original}} > 0.95$ and $\hat{j}$ corresponding precisely to the initial token in all cases.
The probability distributions consistently exhibited sharp peaks at original tokens, demonstrating that optimization occurs within constrained semantic neighborhoods rather than reconstruct the original semantic foundation.

This localized optimization behavior provides crucial advantages over discrete prompt engineering approaches. The continuous embedding space enables sub-token adjustments impossible in discrete text spaces, allowing for precise ``semantic interpolation'' between related concepts—such as blending attributes of ``analyze'' with nuances of ``evaluate''—that gradually shift the prompt's effectiveness without altering its core meaning. This granular control represents a fundamental advance over discrete methods like AutoPrompt~\cite{autoprompt}, which often produce semantically divergent prompts when navigating the combinatorial explosion of token substitutions.

The semantic anchoring effect ensures optimization occurs within a trust region around the original embedding, preventing catastrophic semantic drift while permitting nuanced refinement. This characteristic is particularly valuable for sensitive applications where prompt integrity is essential, such as medical diagnosis or legal analysis, where minor semantic shifts could significantly alter output interpretations. By operating in the continuous embedding space, EmbedGrad achieves what discrete optimization cannot: nuanced adjustments to prompt semantics while preserving the crucial link between human intention and model behavior, effectively bridging the precision of parameter tuning with the interpretability of natural language prompting. The constrained optimization pathway also explains EmbedGrad's consistent performance improvements across diverse tasks without unpredictable semantic jumps that could degrade reliability in production environments.

\subsection{Confidence Enhancement through Entropy Reduction}
Our investigation into EmbedGrad's effectiveness reveals significant changes in model behavior during generation, particularly through the lens of trajectory entropy—a well-established measure of decision confidence in autoregressive models~\cite{entropy_lunch}. 
We quantify this key metric using a rigorous computational approach: for each generated sequence of length $T$, the trajectory entropy is formally defined as:
\begin{equation}
    H_{\text{trajectory}} = -\frac{1}{T}\sum_{t=1}^{T}\sum_{\omega \in \mathcal{V}} p_t(\omega)\log_2p_t(\omega),
\end{equation}
where $p_t(\omega)$ represents the probability distribution over the vocabulary $\mathcal{V}$ at generation step $t$.
Prior to optimization, both Qwen2.5-Math-1.5B and 7B models exhibited problematic generation patterns during mathematical reasoning tasks. Most notably, they frequently entered repetitive output loops, persistently recycling their final phrase until reaching token limits—a failure mode characterized by abnormally low entropy values averaging 0.05. This behavior was especially pronounced in the smaller 1.5B model, occurring in over two-thirds of failed solutions. When models avoided this trap, baseline entropy measured 0.237 for the 1.5B model and 0.124 for the 7B variant, indicating higher uncertainty in the smaller model's reasoning process.

After applying EmbedGrad, we observed transformative improvements: repetitive failures nearly disappeared while trajectory entropy stabilized around 0.120 for both models. This dual effect—eliminating pathological low-entropy states while reducing normal-state entropy—signals a fundamental shift toward more confident and stable generation. For the 1.5B model, this represented a substantial reduction in uncertainty (from 0.237 to 0.120), directly corresponding to its dramatic 44.22\% accuracy improvement on Math500. The 7B model's smaller entropy decrease (from 0.124 to 0.120) aligns with its more modest 9.16\% accuracy gain, demonstrating how improvement magnitude correlates with initial uncertainty levels.

These entropy patterns illuminate EmbedGrad's core mechanism: by refining prompt embeddings, the method provides better initial guidance that streamlines the model's reasoning pathway. This reduces computational hesitaton during generation, where each step requires less probabilistic reevaluation. The convergence of both models' entropy near 0.120 suggests this value represents an optimal confidence zone for reliable mathematical reasoning: sufficiently certain to avoid indecision errors but not so overconfident as to trigger degenerative loops. Critically, the greater entropy reduction in smaller models explains why they benefit more dramatically, as EmbedGrad effectively compensates for their inherent architectural limitations. This positions prompt embedding optimization not merely as an input adjustment, but as a calibration technique that enhances the fundamental confidence-uncertainty balance governing model reasoning.

\subsection{Enhanced Functional Activation via Representation Engineering}
Our analysis extends beyond entropy metrics to reveal how EmbedGrad-optimized embeddings fundamentally reshape model cognition through Representation Engineering. Using Linear Artificial Tomography (LAT)~\cite{representation_engineering}, we quantitatively measure how optimized embeddings activate target functionalities like sentiment classification. This top-down approach characterizes high-level model behaviors by comparing internal representations under two conditions: original text prompts versus optimized embeddings.

\begin{figure}[htbp]
\centering
\includegraphics[width=\linewidth]{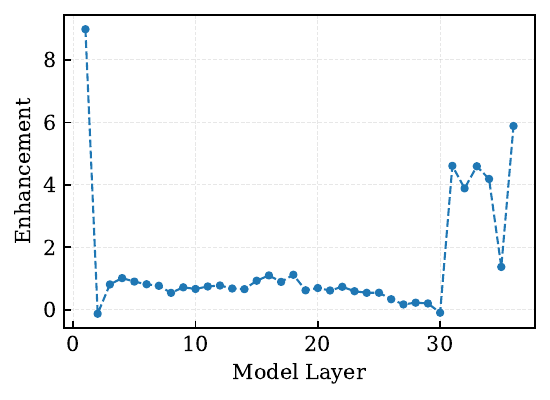}
\caption{Activation intensity enhancement of model's sentiment classification functionality on the optimized input vector across model's layers.}
\label{fig_re}
\end{figure}

The LAT methodology identifies a ``sentiment direction'' in representation space by contrasting emotional and robotic responses to stimuli. Projecting layer-wise activations onto this direction reveals functional activation intensity. As depicted in Figure~\ref{fig_re}, for sentiment classification, optimized embeddings consistently show stronger activation than original prompts across all layers ($\Delta > 0$), with the most dramatic improvement at the first layer. This first-layer amplification establishes a more effective foundation for downstream processing, while subsequent layers maintain moderately elevated activation.

These findings provide mechanistic evidence for EmbedGrad's effectiveness: optimized embeddings activate task-relevant representations more efficiently, particularly at critical input processing stages. The layered enhancement pattern explains performance gains observed in sentiment analysis, where optimized prompts improved accuracy by up to 74.54\% for smaller models. By enhancing functional activation while preserving semantic intent, EmbedGrad demonstrates how continuous embedding refinement unlocks latent model capabilities inaccessible through discrete text modifications.

\subsection{Hyperparameters Sensitivity and Model-Scale Optimization}
Our experimental findings reveal that EmbedGrad's effectiveness exhibits significant sensitivity to optimization hyperparameters, particularly learning rate and iteration count, with improper settings leading to overfitting and suboptimal performance. This sensitivity displays a clear correlation with model scale, as evidenced by our MATH500 experiments: the optimal learning rate for the 1.5B model was approximately 0.01, while the larger 7B model required a more conservative 0.001 rate. Similarly, iteration counts between 5-10 epochs consistently produced the best results across models, with smaller models tolerating slightly higher iteration counts before overfitting manifested.

This model-scale dependency directly connects to our earlier findings about semantic anchoring and entropy reduction mechanisms. Smaller models like the 1.5B variant possess higher baseline uncertainty (evidenced by their 0.237 trajectory entropy), creating a broader optimization landscape that permits more aggressive learning rates without immediate overfitting risks. Their shallower architecture also allows prompt embedding updates to propagate more directly through fewer layers, enabling faster convergence. Conversely, larger models like the 7B version operate with lower baseline entropy (0.124) and more complex internal representations, making their optimization landscape more intricate and prone to overshooting optimal regions with overly ambitious learning rates.

The limited 5-10 epoch sweet spot emerges from the fundamental nature of embedding space optimization. Unlike conventional fine-tuning that adjusts millions of parameters, EmbedGrad optimizes only the prompt embedding vector—a highly compressed target where changes rapidly saturate semantic meaning. Extended iterations risk over-specializing the prompt to the training examples, compromising generalization. This effect mirrors our semantic anchoring observations: optimization occurs within a confined neighborhood of the original embedding, meaning excessive adjustments can breach the trust region boundary, causing semantic drift rather than refinement. 

\subsection{Limitations and Practical Considerations} 
While EmbedGrad demonstrates significant advantages in prompt optimization, we acknowledge several practical limitations that merit consideration. The most notable constraint is its dependence on model accessibility—since the method operates through gradient updates in the embedding space, it requires access to the model's internal architecture and embedding layer, limiting its applicability to open-source models rather than closed API-based systems. This technical requirement excludes deployment scenarios involving proprietary models like GPT-4 or Claude, where only input-output interactions are permitted. Additionally, the optimization's effectiveness remains inherently constrained by the quality of the initial prompt, as the semantic anchoring mechanism confines adjustments within the neighborhood of the starting embedding; fundamentally flawed prompts may require human revision before optimization can yield substantial benefits. The method also exhibits sensitivity to hyperparameter selection, particularly learning rate and iteration count, necessitating careful calibration—typically 3-5 validation runs—to avoid suboptimal performance.

Despite these limitations, EmbedGrad maintains substantial value within its operational domain. For the vast ecosystem of open-source models (LLaMA~\cite{llama}, Qwen~\cite{qwen2.5}, Mistral~\cite{mistral}, etc.), it provides a uniquely efficient optimization pathway. The semantic preservation characteristic is actually advantageous in safety-critical applications like healthcare and finance, where prompt integrity is paramount, and its limitations become strengths by preventing uncontrolled semantic drift.  Future work will focus on developing transfer learning techniques to extend optimized embeddings across model families and creating distillation methods that allow closed-model deployment. These practical constraints do not diminish EmbedGrad's core innovation—its ability to bridge discrete prompt engineering and continuous parameter optimization—but rather define the boundary conditions where its transformative potential is maximized: enabling resource-efficient, interpretable, and robust prompt enhancement for the open-source AI ecosystem.

\section{Conclusion and future work}\label{sec:con}
EmbedGrad pioneers a gradient-based approach to prompt optimization by refining prompt embeddings within continuous semantic space, fundamentally advancing beyond traditional discrete text modifications. Our method achieves precise prompt calibration while maintaining human interpretability and computational efficiency through semantic anchoring—preserving core meaning while enabling nuanced adjustments. Comprehensive evaluations across mathematical reasoning and sentiment analysis demonstrate significant performance enhancements, particularly benefiting smaller models by compensating for architectural limitations. Although currently optimized for open-source implementations and dependent on initial prompt quality, EmbedGrad establishes a new paradigm for parameter-efficient adaptation that bridges human intention with machine reasoning. Future work will extend this framework to cross-model transfer learning and distillation techniques, further enhancing accessibility while preserving the crucial balance between precision and interpretability in AI systems.

\bibliography{aaai2026}

\end{document}